# Virtual Coil Augmentation Technology for MR Coil Extrapolation via Deep Learning

Cailian Yang, Xianghao Liao, Yuhao Wang, *Senior Member, IEEE*, Minghui Zhang, Qiegen Liu, *Senior Member, IEEE*

*Abstract*—Magnetic resonance imaging (MRI) is a widely used medical imaging modality. However, due to the limitations in hardware, scan time, and throughput, it is often clinically challenging to obtain high-quality MR images. In this article, we propose a method of using artificial intelligence to expand the channel to achieve the goal of generating the virtual coils. The main characteristic of our work is utilizing dummy variable technology to expand/extrapolate the receive coils in both image and k-space domains. The high-dimensional information formed by channel expansion is used as the prior information to improve the reconstruction effect of parallel imaging. Two main components are incorporated into the network design, namely variable augmentation technology and sum of squares (SOS) objective function. Variable augmentation provides the network with more high-dimensional prior information, which is helpful for the network to extract the deep feature information of the data. The SOS objective function is employed to solve the deficiency of k-space data training while speeding up convergence. Experimental results demonstrated its great potentials in super-resolution of MR images and accelerated parallel imaging reconstruction.

## I. Introduction

FROM the development of the magnetic resonance imaging (MRI) phased array until recently, most MRI scanners employed a modest number of receiver coils which are used primarily to enhance the signal-to-noise ratio (SNR) over relatively large fields of view (FOV). The simultaneous use of multiple receiver coils was pioneered in the late 1980s [1]-[3] and analyzed in detail by Roemer *et al.* [4]. Due to the emergence of parallel imaging (PI) technology [5]-[9], the number of MRI receive coils have recently blossomed. A number of groups have reported the experimental results from 32 coils [5] to 64 coils [6], with recent conferences have reported presenting preliminary work on 90 coils [7], 96 coils [8], and 128 coils [9]. The simultaneous use of a large number of receiver coils in principle enables high acceleration factors for PI and improves SNR.

Due to the obvious benefits of large coil arrays for SNR and acceleration, both researchers and manufacturers have been developing and building large dedicated arrays [10]-[13]. However, packing receiver-coil electronics into a small region can compound coil coupling and losses, affect flexibility and weight. Additionally, it causes increased memory usage and large computational costs for reconstructing the missing data from such a large number of coils. In this study, we will develop artificial intelligence (AI)-assisted channel expansion/extrapolate technology in PI to enhance spatial resolution and fast reconstruction. For the convenience of further description, we list some background of super-resolution (SR) in image domain and PI in k-space domain, respectively.

### A. Review of SR

Roughly speaking, SR algorithms of MR images can be divided into two categories: Model-based and learning-based approaches. Interpolation algorithms [14]-[16], like bilinear, bicubic, and nearest-neighbor interpolation techniques, are representatives of simple model-based approaches, which can be directly used to enlarge images. Wiener filtering and iterative deblurring algorithms utilize knowledge on point spread function of the system to recover image resolution [17], [18] are also considered as model-based algorithms. Dictionary learning-based SR techniques [19], [20] are examples of learning-based algorithms. The performance of these conventional methods is essentially limited, since they apply inadequate additional information and models with limited representational capacity to solve the notoriously challenging ill-posed inverse problem of SR tasks.

Over the past decades, deep learning methods have expanded the learning-based SR methods and gained increasing attention. Especially, convolutional neural networks (CNN) based models have become one kind of the most popular methods in SR of MR images, because of their simple network structure and high restoration accuracy. As a pioneer work, Dong *et al.* [21] first proposed the convolutional CNN-based SR (SRCNN) algorithm, which directly learned an end-to-end mapping from LR images to HR images. Pham *et al.* [22] developed a 3D SRCNN algorithm for the SR of Brain MR images. Besides, Chen *et al.* [23] provided a trusted deep CNN-based SR method for HR medical image reconstruction, which can transmit the information of the output image to the low-level features by a feedback connection. Shi *et al.* [24] proposed a residual learning-based SR algorithm (MGLRL, multi-scale global-local residual learning), which integrated both the multi-scale global residual learning and local residual learning. McDonagh *et al.* [25] developed a context-sensitive SR algorithm with the residual CNN that learned organ specific appearances and generated HR images with sharp edges and rich details. SRGAN [26], which was developed based on GAN, had demonstrated perceptually better results compared to other deep residuals. Furthermore, Chen *et al.* [27] invented a multi-level densely connected medical image SR network (mDCSRN). Zhu *et al.* [28] proposed a lesion focused SR (LFSR) method, which incorporates GAN to

This work was supported by National Natural Science Foundation of China (61871206, 61601450).

C. Yang, X. Liao, Y. Wang, M. Zhang, and Q. Liu are with the Department of Electronic Information Engineering, Nanchang University, Nanchang 330031, China. ({yangcailian, liaoxianghao}@email.ncu.edu.cn, {wangyuhao, zhangminghui, liuqiegen}@ncu.edu.cn).

achieve SR results for MR images. Although there were plenty of progresses in SR performance, all of them are implemented in spatial domain. Therefore, the challenge remains for achieving superior SR on MR images in channel domain.

*B. Review of PI*

PI is a robust way to accelerate data acquisition. The basis of PI is the simultaneous and independent data acquisition using receiver arrays with multiple radiofrequency coils. Scan time reductions are achieved by reducing the amount of gradient encoding steps normally required to obtain an artifact-free image. Reduced gradient encoding leads to under-sampling of k-space, resulting in aliasing artifacts. Dedicated PI algorithms such as simultaneous acquisition of spatial harmonics (SMASH) [29], sensitivity encoding (SENSE) [30], or generalized auto-calibrating partially parallel acquisitions (GRAPPA) [31] take advantage of spatial coil sensitivity variations to remove the aliasing artifacts.

The popular SENSE [30] approach is a pre-calibrated scheme that unfolds superimposed pixels by incorporating spatial coil sensitivity information. However, a challenge in calibration-based methods is the potential for motion artifacts, which results from mismatches between the calibration and main scans. In GRAPPA [31], missing k-space lines in a single-coil are approximated by a linear combination of measured k-space lines from all coils. To calculate the reconstruction coefficients, additional Nyquist sampled k-space lines (auto-calibration signal, ACS) have to be measured. Recently, deep learning methods RAKI [32] and LO-RAKI [33] that learn neural networks from the auto-calibration region have also been applied to improve the image quality at high acceleration factors. Despite these advances, the need to acquire fully sampled auto-calibration regions often restricts the achievable acceleration and the amount of higher k-space samples that can be acquired in a realistic scan time. To minimize these trade-offs, several researchers have exploited calibration-free approaches. Typical examples include simultaneous auto-calibrating and k-space estimation (SAKE) [34] as well as P-LORAKS [35]. These techniques do not need specific calibration regions/pre-scans to estimate the interpolation kernels or sensitivities.

*C. Contribution*

In this article, we aim to introduce a method of using artificial intelligence to virtually expand/extrapolate the coil arrays in PI via virtual coil augmentation technology, namely VCA. Specifically, we use dummy variable technology to expand the channel/coil to achieve the purpose of increasing the virtual coil arrays. The main characteristic is the use of dummy variable technology to expand the channel in both image and k-space domains, respectively. The resulting algorithms are coined VCA-I and VCA-K, accordingly. Although both of them are operated under a unified network framework, there exist essential differences between them. The input and output of VCA-I are the data in image domain, while the input and output of VCA-K are directly from the measure data in k-space. The sum of squares (SOS) objective function in both VCA-I and VCA-K are executed in image domain. Among them, the preliminary results in image domain have been already published in the ICMIPE conference [36] and an oral presentation has been made. In this study, we have carried out substantial advances on the preliminary work.

The main contributions of this work are summarized as follows:
- Channel expansion/extrapolation is performed in image domain and k-space domain, respectively. The high-dimensional information formed by VCA-I is used as the prior information to leverage the SR of MR images, while the counterpart formed by VCA-K is devoted to accelerating PI such as to improve the reconstruction effect.
- Using the reversible network as the training tool, auxiliary variables technology is adopted to make the channel number of the network output to be equal to the network input. The variable augmentation technology provides higher-dimensional prior information for the network, which is helpful for the network to extract the deep feature information of the data.
- The SOS objective function is designed to handle the deficiency of k-space data training and accelerate convergence. It is worth noting that the input and output of VCA-K in training procedure are both from k-space domain, while the SOS objective function is performed in image domain.

## II. PRELIMINARY

*A. PI Reconstruction*

Before discussing the details of the multiple coils, we first review some basics of PI technique. The formation of aliased images from multiple receivers in PI can be formulated as a linear operation to "fold" the full-FOV spin density images, i.e.,

$$y = Ax \qquad (1)$$

where $y$ is the vector formed from the pixel intensities recorded by each receiver (folded image), and $x$ is the vector formed from the full-FOV image. The encoding matrix $A$ consists of the product of the aliasing operation due to subsampling of the k-space data and coil-specific sensitivity modulation over the image. The goal of the image reconstruction is to solve for $x$ given our knowledge of $A$. In general, Eq. (1) is an over-determined linear system, i.e., the number of array coils, which is the row dimension of $y$, exceeds the number of the pixels that fold into the measured pixel, the row dimension of $x$. If the matrix is well conditioned, the inversion can be achieved with minimal amplification of noise. While the encoding matrix can still be inverted even if it is nearly singular, in this ill-conditioned case, small noise perturbations in the measured data (aliased image) can produce large variations in the full-FOV reconstruction. This effect causes noise amplifications in regions of the image where the encoding matrix is ill-conditioned.

To date, SENSE and GRAPPA remain the most common partially PI techniques. Both of these techniques increase imaging speed by reducing the number of phase encoding steps to encode an image. In SENSE, a set of $N$ aliased images are acquired from an $N$ element receive array. The sensitivity patterns for each element in the array are adopted to form a set of linear equations which relate the aliased images from each coil to the unaliased full field-of-view image. Solving this set of equations performs an "unwrapping" of the data in image domain to yield the desired image. In contrast to SENSE, GRAPPA is not iterative and does not require prior knowledge of coil sensitivity profiles. Recon-struction is performed directly in k-space by estimating missing data points as weighted linear combinations of known acquired points. The GRAPPA weights implicitly contain information about the coil sensitivity profiles and are determined using one or more fully sampled training scans. Aliasing artifacts are caused by insufficient phase-encoding line sampling in k-space, while linear convolution kernels are susceptible to the noisy at high acceleration factors. Coil sensitivities are most similar in the ACS region and are more susceptible to noise. Increasing the number of coils can reduce the image support, thereby alleviating the ill-posedness of image reconstruction and improving image quality.

*B. Multiple Coils in Hardware*

The idea of using an array of multiple and independent receive coils to reduce MR imaging time was first introduced in 1987 by two different groups [37], [38]. Indeed, Hutchinson and Raff went so far as to claim that a full $N \times M$ image could be acquired entirely without phase encoding by placing $N$ closely spaced small receive coils around a sample [39]. The main deficiency of this completely PI method is that receiving an image from $N$ receive coils requires a receiver with at least $N$ channels.

The idea of using multiple coils for imaging reconstruction has been the subject of some interest in MRI community. The radio frequency coil of the system presented in [39] consists of four independent current sheet antennas. A head coil consisting of eight overlapping loop elements was used for imaging in [40]. A degenerated birdcage resonator according to [41] for whole-body imaging was presented. Furthermore, Vernicke *et al.* [42] described the development of a multielement body coil with eight transmission electron microscope elements for application in a fully integrated multichannel MRI system. Massner *et al.* [43] proposed the concept of mechanically adjustable MR receiver coil arrays. An 8-channel wrist array for proton MRI at 3.0 T was constructed and evaluated. The array is adjusted to the individual anatomy by a mechanism that fitted a configuration of flexible coil elements closely around the wrist. Thalhammer *et al.* [44] introduced a 2D 16-channel receive coil array, which is tailored for cardiac magnetic resonance imaging at 7.0 T. The coil array consists of two sections, each using eight elements arranged in an $2 \times 4$ array. Graessl *et al.* [5] designed a modular transceiver coil array with 32 independent channels and received coil array for Cardiac MRI at 7.0 T. The modular coil array comprises eight independent building blocks, each containing four transceiver loop elements. Keil *et al.* [6] designed a 64-channel Brain array coil. The posterior former of the 64-channel array incorporated 39 hexagons and three pentagons. The anterior section contains 15 hexagons, three pentagons, two noncircular-shaped loops at the edge of the housing, and two somewhat oval-shaped eye loops.

These researches have shown that multiple coils have the potential for increasing throughput of the MR imaging experiment via parallelization. Until recently, the number of independent receiver channel supported by MR systems was limited to typically 32 to 64. New spectrometer hardware is becoming available that permits 96 and more receiver channels to be used, which has triggered research into applying large coil arrays with 128 independent receive channels. Fig. 1 visualizes an 8-channel independent receiver MR system. However, for large coil counts this is tedious to do in practice and requires cumbersome means of geometric fixation, which will lead to uncomfortable for the patients.

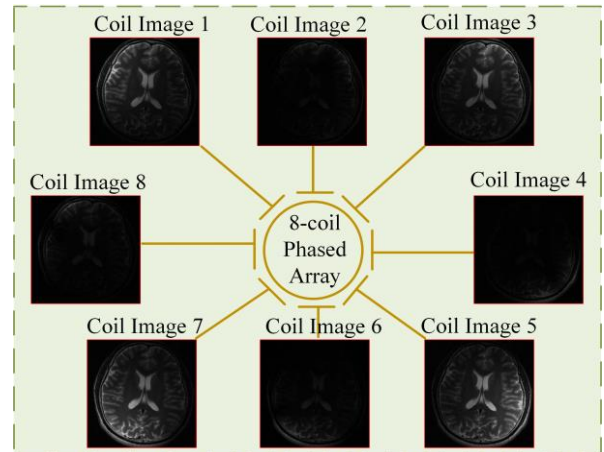

**Fig. 1.** An image obtained from an 8-coil phased array. Notably, each coil is more sensitive to the signal from the tissue closest to it and can be used to form its image (small peripheral image). Independent coil images can be combined into a single image with uniform sensitivity.

*C. Virtually Multiple Coils*

Large physical coil arrays enable faster data acquisition with high SNR. Nevertheless, the construction and data storage of these large datasets are becoming increasingly problematic. To further improve the effect of PI, some researchers have proposed the concept of virtual coils [45], [46]. Generating additional data in the form of virtual coils is an efficient way to improve PI reconstruction. For example, Liao *et al.* [47] proposed a virtual coil (VC) reconstruction framework to improve highly accelerated single-shot echo planer imaging (EPI) and generalized slice dithered enhanced resolution acquisition in high-resolution diffusion imaging. However, VC requires the phase to be consistent between the reference data and imaging data. Blaimer *et al.* [45] presented a new approach to improve PI performance by utilizing conjugate k-space symmetry. By generating virtual coils containing conjugate symmetric k-space signals from actual coils, additional image-phase and coil-phase information can be incorporated into the reconstruction process for parallel acquisition techniques. Besides, VCC-GRAPPA [48] combines the virtual conjugate coil concept with k-space algorithms GRAPPA for image reconstruction. Extra spatial encoding capability provided by the phase map of VCC would improve GRAPPA inversing conditions. In addition, Wang *et al.* proposed NL-VCC-GRAPPA [49] to jointly utilize the nonlinear mapped virtual coil and phase conjugated virtual coil to further reduce noise amplification in PI. The idea of virtual coil has been generalized by Bilgic *et al.* [50] from VCC to further utilize information redundancy from the multi-contrast, multi-echo, or phase cycled BSSFP datasets to improve the PI reconstruction condition. Very recently, Liu *et al.* [51] combined single-shot variant EPI (ssDW-EPI) with virtual coil acquisition and developed a self-reference reconstruction method to eliminate the residual aliasing artifact on multi-oblique ssDW-EPI sequence with PI and multiple signals averaging.

Although some attempts have been made for generating virtual coils, the number of additional coils has been greatly limited, and the correlation between the generated virtual coils and the original coils is relatively large. For example, VCC-GRAPPA can only achieve the effect of double expansion. Due to the recent development of deep learning tech-nology in imaging, it is urgent to utilize deep learning method to generate a large number of virtual coils. Fig. 2 demonstrates a rough schematic of using deep learning method to generate virtual coils. Specifically, the independent receiving system collects 4-channel data. By means of deep learning, it can map to an object with 8-channel or even more channels. Thus, the advantage of exploiting large number of coils can be utilized.

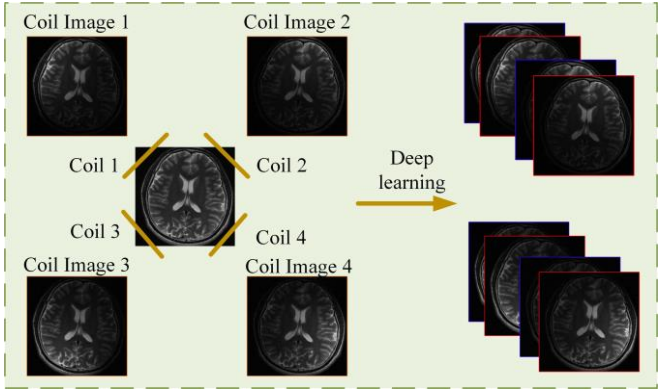

**Fig. 2.** A diagrammatic sketch of the virtual coil augmentation technology. The images in the red-square boxes belongs to those generated by the virtual coil technique.

## III. PROPOSED VCA MODEL

In the previous section, we discussed the basic concepts of PI, multi-coils in hardware surroundings, and virtual multi-coils. In this section, we mainly introduce the proposed channel expansion/extrapolation method in the version of image domain and k-space domain, respectively. The network architecture and loss function of the proposed method are also presented in this part. Particularly, this article uses variable augmentation technology to expand/extrapolate the channel in image domain and k-space domain to tackle the deficiency of low image resolution and slow imaging speed, respectively.

### A. Proposed VCA

As well known, MR image is different from natural image, which is a complex-valued image. In this study, if the 12-channel $256 \times 256$ complex-valued images $x$ are used as the training and testing data in our model, each channel with complex-valued can be represented by 2-channel with real value containing real and imaginary components. Thus, we acquire 24-channel $256 \times 256$ input images $X$. In other words, $X = \{[x_{real1}, x_{imag1}], \cdots, [x_{real\,n}, x_{imag\,n}]\}$ is used as the input of the network.

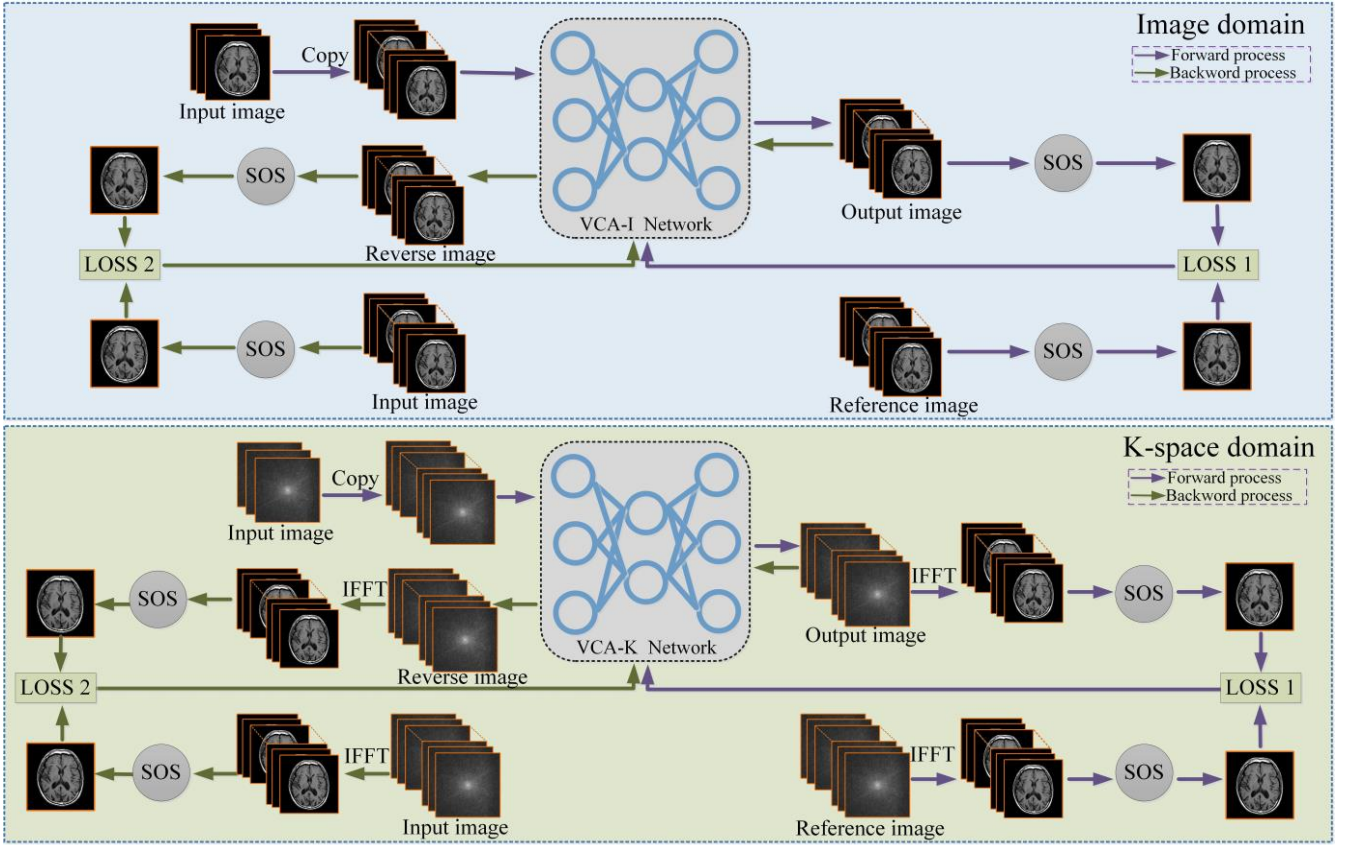

**Fig. 3.** The training pipeline of VCA. Top line: VCA-I that conducts the channel expansion in image domain; Bottom line: VCA-K that conducts the channel expansion in k-space domain. Among them, variable augmentation technology is used in the stage of image pre-processing. The loss function is composed of a forward loss function and a reverse loss function. For the convenience of readers, the detailed architecture of VCA is described in Section III. B. On the other hand, the description of the loss functions in the training objectives is provided in Section III. C.

Reversible network is selected as the central component of VCA for training and estimation. Specifically, the model is designed with the composition of a stack of affine coupling layers and utilizes the invertible $1 \times 1$ convolution as the learnable permutation function between the coupling layers. Note that the channel number of network input and output must be equal, due to the invertible convolution and connection operations. Inspired by the work in [52], in this study we use dummy variable technology to increase the channel dimension to make the number of input and output channels of the reversible network to be consistent. Specifically, taking the example of extrapolating 2 coils object to be 12 coils counterpart for instance, we copy 2-channel complex-valued image into 12-channel, which is stacked to be 24-channel as the input of the network, and the output of the network is also 24-channel. Finally, the reconstructed images can be obtained after SOS operation.

During the training stage, the difference between reconstructed images and ground truth images is not only applied to adjust reconstruction layer to restore better images from manifold, but also to guide extraction of accurate image features. The top part of Fig. 3 visualizes the channel expansion method in image domain, and the bottom part depicts the channel expansion method in k-space domain. Among them, variable enhancement technology is utilized in the process of image pre-processing. A combination of the forward loss and backward loss is employed to optimize VCA.

Although both VCA-I and VCA-K are carried out under a unified network framework, there exist distinct differences between them. More precisely, the input and output of VCA-I are the data of image domain, while the input and output of VCA-K are the data of k-space domain. The SOS objective function of both VCA-I and VCA-K are executed in image domain. As well known, the k-space samples are in spatial frequency domain. Because the amplitude difference in the magnetic resonance k-space is too large, it is impossible to train a better model by directly calculating the objective function in frequency domain. VCA-K calculates the objective function in image domain, so that the features can be extracted well in the network training process. At the same time, the SOS is used to further improve the network performance.

*B. Network Architecture*

Our goal is to find a bijective function which can map the data point from input data space $X$ to output data space $Y$. To achieve this, classical neural networks need two separate networks to approximate $X \to Y$ and $Y \to X$ mappings respectively, which leads to inaccurate bijective mapping and may accumulate the error of one mapping into the other. We take an alternative method and use the affine coupling layers in [53], [54] to enable the invertibility of one single network.

We design VCA with the composition of a stack of invertible and tractable bijective functions $\{f_i\}_{i=0}^k$, i.e., $f = f_0 \circ f_1 \circ f_2 \circ \cdots \circ f_k$. Inputting the image $x$ into the invertible network $f$ yields the output image $y$ and vice versa. Thus, the information is fully preserved during both the forward and reverse transformations. For a given observed data sample $x$, we can derive the transformation to target data sample $y$ through:

$$y = f_0 \circ f_1 \circ f_2 \circ \cdots \circ f_k(x) \quad (2)$$
$$x = f_k^{-1} \circ f_{k-1}^{-1} \circ \cdots \circ f_0^{-1}(y) \quad (3)$$

The bijective model $f_i$ is implemented through affine coupling layers. The affine coupling layer provides a powerful reversible transformation where the forward function, the reverse function and the log-determinant are computationally efficient. In each affine coupling layer, given a $D$ dimensional input $m$ and $d < D$, the output $n$ is calculated as:

$$n_{1:d} = m_{1:d} \quad (4)$$
$$n_{d+1:D} = m_{d+1:D} \odot \exp(s(m_{1:d})) + t(m_{1:d}) \quad (5)$$

where $s$ and $t$ represent scale and translation functions from $R^d \to R^{D-d}$, and $\odot$ is the Hadamard product. Note that the scale and translation functions are not necessarily invertible, and thus they are realized by a succession of several convolution layers with leaky ReLU activations.

As stated in [54], the coupling layer leaves some input channels unchanged, which greatly restricts the representation learning power of this architecture. To alleviate this deficiency, we firstly enhance [55] the coupling layer by:

$$n_{1:d} = m_{1:d} + r(m_{d+1:D}) \quad (6)$$

where $r$ can be an arbitrary function from $R^{D-d} \to R^d$. The inverse step is easily obtained by:

$$m_{d+1:D} = (n_{d+1:D} - t(n_{1:D})) \odot \exp(-s(n_{1:d})) \quad (7)$$
$$m_{1:d} = n_{1:d} - r(m_{d+1:D}) \quad (8)$$

Each step of flow should be preceded by some kind of permutation of the variables to ensure that each dimension can affect every other dimension after sufficient steps of flow. Next, we utilize the invertible 1×1 convolution proposed in [53] as the learnable permutation function to reverse the order of channels for the next affine coupling layer. The weight matrix is initialized as a random rotation matrix. Note that a 1×1 convolution with the equal number of input and output channels is a generalization of the permutation operation.

The elaborate architecture of VCA is shown in Fig. 4. It contains several invertible blocks, where each invertible block consists of invertible 1×1 convolution and affine coupling layers. The split function splits the input tensor into two halves along the channel dimension, while the concat operation performs the corresponding reverse operation. In [56], another type of split was introduced that is along the spatial dimensions using a checkerboard pattern. In this work we only perform splits along the channel dimension, simplifying the overall architecture. $s$, $t$ and $r$ are transformation equal to dense block, which consists of five 2D convolution layers with filter size 3×3. Each layer learns a new set of feature maps from the previous layer. The size of the receptive field for the first four convolutional layers is 3×3, and stride is 2, followed by a rectified linear unit (ReLU). The last layer is a 3×3 convolution without ReLU. The purpose of the Leaky ReLU layers is to avoid overfitting to the training set [57] and further increase nonlinearity. Invertible model is composed of both forward and inverse processes. During the training time, the input image is transformed to reconstructed output image by a stack of bijective functions $\{f_i\}_{i=0}^k$.

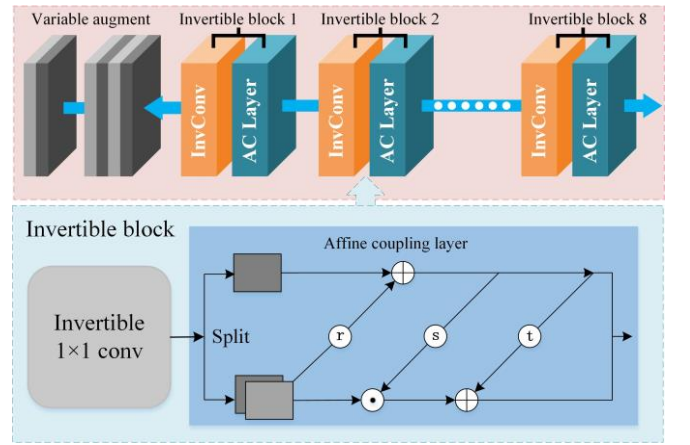

**Fig. 4.** The pipeline of VCA. Invertible model is composed of both forward and inverse processes. We illustrate the details of the invertible block on the bottom. $r$, $s$, and $t$ are transformations defined in the bijective functions $\{f_i\}_{i=0}^k$.

## C. Training Objectives

For obtaining faithful and satisfactory estimation, we conduct bi-directional training with smooth $L_1$ loss. The gradient of smooth $L_1$ loss function is constant when $x$ is large, which makes up the deficiency that large gradient in $L_2$ loss will destroy training parameters. When $x$ is small, the gradient will dynamically decrease, which can improve the speed of convergence in $L_1$ loss. It is worth noting that although VCA-I and VCA-K are performed in image and k-space domain individually, the SOS and loss functions of both VCA-I and VCA-K are performed in image domain. Therefore, the network output image and label image of VCA-K need inverse Fourier transform before calculating the loss function.

The bi-directional loss function consists of forward loss $L_f$ and backward loss $L_r$. VCA model optimizes the total loss $L$ by conducting joint training on the following two objectives:

$$L = \alpha L_f + \beta L_r \qquad (9)$$

where $\alpha$ and $\beta$ are coefficients for balancing different loss terms. The smooth $L_1$ is defined as:

$$smooth_{L_1}(d) = \begin{cases} 0.5d^2 & if \ |d| < 1 \\ |d| - 0.5 & otherwise \end{cases} \qquad (10)$$

where $d$ represents the error between predicted value and ground truth. The forward process can be expressed as:

$$L_f = \left\| SOS(x) - SOS(x_f) \right\|_1^{smooth} \qquad (11)$$

where $x_f$ is the output of variable augmented invertible network, and $x$ denotes the label image. SOS is defined as $SOS(x_f) = (\sum_{k=1}^{N} |x_f|^2)^{1/2}$, and $x_f$ is the complex-valued image from channel $k$. The backward process can be expressed as:

$$L_r = \left\| SOS(x) - SOS(x_r) \right\|_1^{smooth} \qquad (12)$$

where $x_r$ denotes the reversed resulting image.

## IV. EXPERIMENTS

In this section, the implementation details of VCA are introduced, as well as the datasets used for evaluation. Subsequently, we not only perform an ablation experiment of VCA-I in image domain and compare it with the SR algorithm, but also conduct ablation experiments of VCA-K in k-space domain and perform PI reconstruction on the dataset after the channel expansion.

### A. Experiment Setup

**Datasets.** We use two datasets to evaluate the performance of VCA, which are provided by Shenzhen Institutes of Advanced Technology, the Chinese Academy of Science. First, the Brain data are scanned from a 3T Siemens's MAGNETOM Trio scanner using the T2-weighted turbo spin echo sequence. 480 MR images with 12-channel are chosen as the training data and 20 images are selected as validation data. The relevant imaging parameters include the size of image acquisition matrix is $256 \times 256$, echo time (TE) is 149 $ms$, and repetition time (TR) is 2500 $ms$. The FOV is $220 \times 220$ $mm^2$ and the slice thickness is 0.86 $mm$. Second, 101 fully-sampled Cardiac MR images are acquired with T1-weighted FLASH sequences via using 3T MRI scanner (Siemens's MAGNETOM Trio), which receiver coils number is 20. We select 21 MR images as the verification dataset. Typically, the TR/TE is 5/3 $ms$, acquisition matrix size is $216 \times 216$, FOV is $330 \times 330$ $mm^2$, and slice thickness is 6 $mm$. Our architecture uses 2D axial-plane slices of the volumes as the network input. For each volume, we linearly scale the original intensity values to [−1, 1].

**Model Training.** All networks are trained using the Adam solver. We conduct 300 epochs to train the proposed model. The initial learning rate is set to 0.0001 for the first 50 epochs. The learning rate is halved after every 50 epochs to achieve optimal results. During the training phase, the trade-off parameter $\alpha$ and $\beta$ are set to 12 and 1. Data augmentation is performed on the training images, which is randomly rotated by 90°, 180°, 270° and flipped horizontally to obtain more training data. The training and testing experiments are performed with a customized version of Pytorch on an Intel i7-6900K CPU and a GeForce Titan XP GPU. For the convenience of reproducible research, source code of VCA can be downloaded from the website: https://github.com/yqx7150/VCA.

**Quality Metrics.** To evaluate the quality of the reconstructed images, two metrics, the popular peak signal to noise ratio (PSNR) and the powerful perceptual quality metrics structural similarity (SSIM) are selected for quantitative assessment.

Specifically, the PSNR describes relationship of the maximum possible power of a signal with the power of noise corruption. Higher PSNR means better image quality. Denoting $x$ and $\hat{x}$ to be the reconstructed image and ground truth, PSNR is expressed as:

$$PSNR(x, \hat{x}) = 20 \log_{10}[Max(\hat{x}) / \|x - \hat{x}\|_2]. \qquad (13)$$

The SSIM is used to measure the similarity between the original MR images and reconstructed images, evaluated on three aspects: Luminance, contrast, and structural correlation. SSIM value is normalized between 0 and 1. SSIM is defined as:

$$SSIM(x, \hat{x}) = \frac{(2\mu_x \mu_{\hat{x}} + c_1)(2\sigma_{x\hat{x}} + c_2)}{(\mu_x^2 + \mu_{\hat{x}}^2 + c_1)(\sigma_x^2 + \sigma_{\hat{x}}^2 + c_2)} \qquad (14)$$

where $\mu_x$ and $\sigma_x^2$ are the average and variance of $x$. $\sigma_{x\hat{x}}$ is the covariance of $x$ and $\hat{x}$. $c_1$ and $c_2$ are used to maintain a stable constant.

### B. Ablation Study in Image Domain

The SCC [58] and GCC [59] algorithm can be used for image compression in image domain as well as in k-space domain. We use the SCC algorithm to compress the 12-channel Brain dataset and the 20-channel Cardiac dataset into 2-channel as input for training. The Brain dataset is compressed to be 4-channel and 6-channel using the GCC algorithm, while the Cardiac dataset is compressed to 6-channel and 10-channel as label images. A higher number of ACS lines with an auto-calibrated signal produces better compressed data, but requires a longer compression time. The number of ACS lines is selected as 24 for image compression in our experiment.

To verify the effect of virtual coils, we copy the 2-channel image into different channels through variable enhancement technology for reconstruction. To show a clearer visual effect, we demonstrate the residual results of the reference image and the reconstructed image in Fig. 5 and Fig. 6. Vis-ually, VCA-I algorithm has achieved good reconstruction results in different channel expansions. In particular, the reconstruction results of 12-channel Brain image and 20-channel Cardiac image are sharpest and closest to the ground truth images. Quantitatively, we calculate SSIM and PSNR metrics between output and ground truth images and report in Table I. VCA-I obtains high results in both two indexes. The average PSNR of VCA-I can reach 39 dB and 25.12 dB on two datasets, respectively. As the number of channels increases, the results of reconstruction after SOS are better. In other words, VCA-I enhances the correlation of data through virtual coils technology, which is conducive to the formation of characterization information, thereby improving the reconstruction results.

TABLE I
QUANTITATIVE RECONSTRUCTION (PSNR/SSIM) RESULTS FOR VCA-I ON BRAIN AND CARDIAC DATASETS.

| Brain | 2ch → 4ch | 2ch → 6ch | 2ch → 12ch |
|---|---|---|---|
| | 37.27/0.9880 | 38.84/0.9869 | 41.00/0.9887 |
| Cardiac | 2ch → 6ch | 2ch → 10ch | 2ch → 20ch |
| | 24.79/0.8814 | 25.12/0.8817 | 25.45/0.8771 |

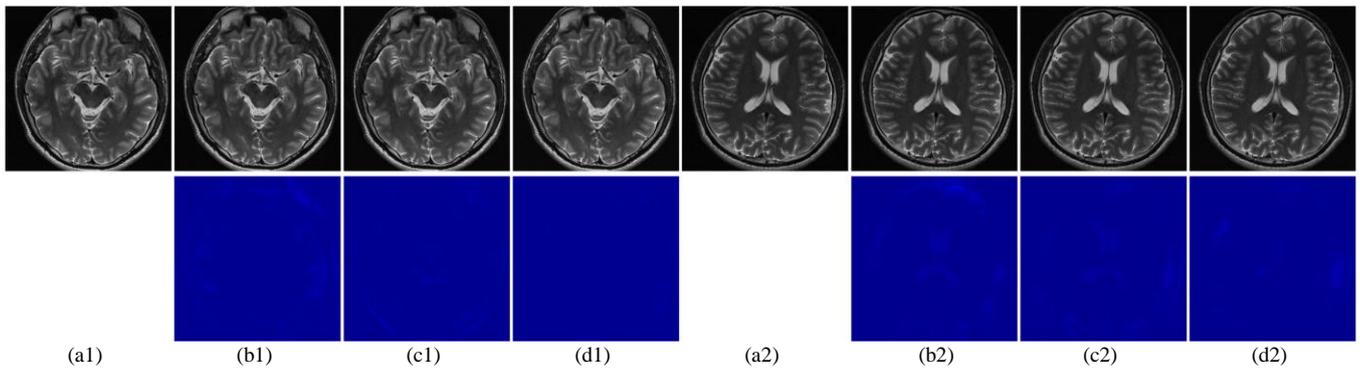

(a1) (b1) (c1) (d1) (a2) (b2) (c2) (d2)
**Fig. 5.** Reconstruction results of Brain dataset for VCA-I. Top: (a) Reference image of Brain dataset; Reconstruction results of (b) 2ch → 4ch; (c) 2ch → 6ch; (d) 2ch → 12ch; Bottom: The residuals between the reference images and reconstruction images.

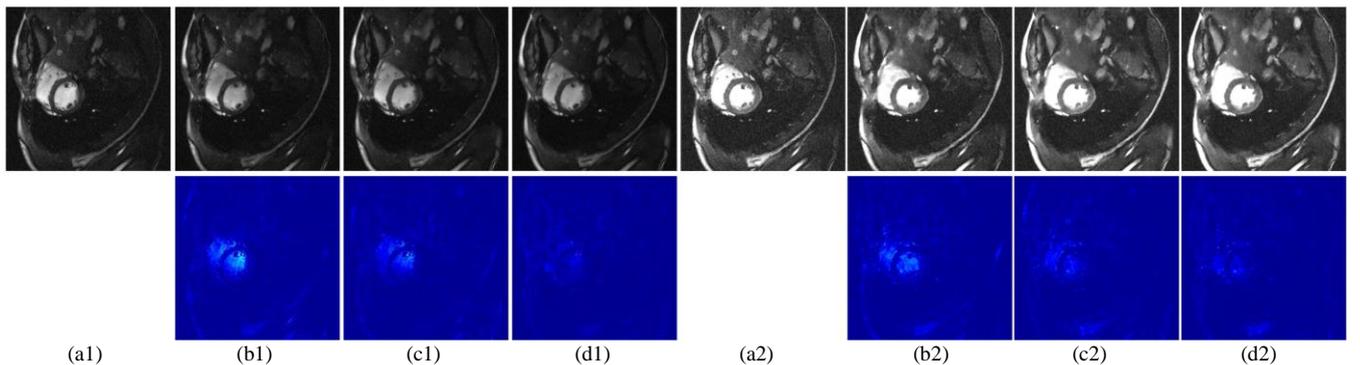

(a1) (b1) (c1) (d1) (a2) (b2) (c2) (d2)
**Fig. 6.** Reconstruction results of Cardiac dataset for VCA-I. Top: (a) Reference image of Cardiac dataset; Reconstruction results of (b) 2ch → 6ch; (c) 2ch → 10ch; (d) 2ch → 20ch; Bottom: The residuals between the reference images and reconstruction images.

*C. SR Comparison in Image Domain*

In this part, we compare VCA-I with four state-of-the-art SR methods. These well-known benchmark SR methods include BICUBIC [16], SRCNN [21], FAWDN [23], and SRGAN [26]. Among them, BICUBIC is a traditional interpolation method. SRCNN is the earliest proposed SR algorithm based on CNN, which directly learns the end-to-end mapping from LR images to HR images. FAWDN is a deep convolutional neural network-based SR method for HR medical image reconstruction. SRGAN is a typical GAN-based SR algorithm. In our experiments, the networks of all the algorithms are retrained for database. Two images with sufficient texture details are selected from the testing set to validate the performance of each SR method. The comparisons between the high-resolution images obtained by each method and the real high-resolution images (Ground Truth, GT) are shown in Fig. 7.

Encouragingly, VCA-I outperforms these methods as evidenced by the resultant SR images with higher PSNR and SSIM values. For instance, VCA-I can reach 41 dB, which is 11.54 dB, 9.1 dB, 6.99 dB, and 8.64 dB higher than BICUBIC, SRCNN, FAWDN, and SRGAN algorithms, respectively. Compared with SRGAN, the most distinct characteristic of VCA-I is that artifacts and noise are greatly reduced. Also, VCA-I shows richer textural details than FAWDN and SRCNN. Since the interpolations do not introduce any new details into the enlarged images, Bicubic easily produces ringing and excessive smoothing. For example, from the residual images in Fig. 5, it is observed that the texture details of VCA-I are clearer and closer to the ground truth, and these textures are blurred in the images generated by other methods.

TABLE II
COMPARISON RESULTS WITH FOUR STATE-OF-THE-ART METHODS ON BRAIN DATASET IN TERMS OF PSNR AND SSIM.

| Dataset | BICUBIC | SRCNN | FAWDN | SRGAN | VCA-I |
|---|---|---|---|---|---|
| Brain | 29.46/0.8967 | 31.90/0.9316 | 34.01/0.9356 | 32.36/0.9354 | **41.00/0.9887** |

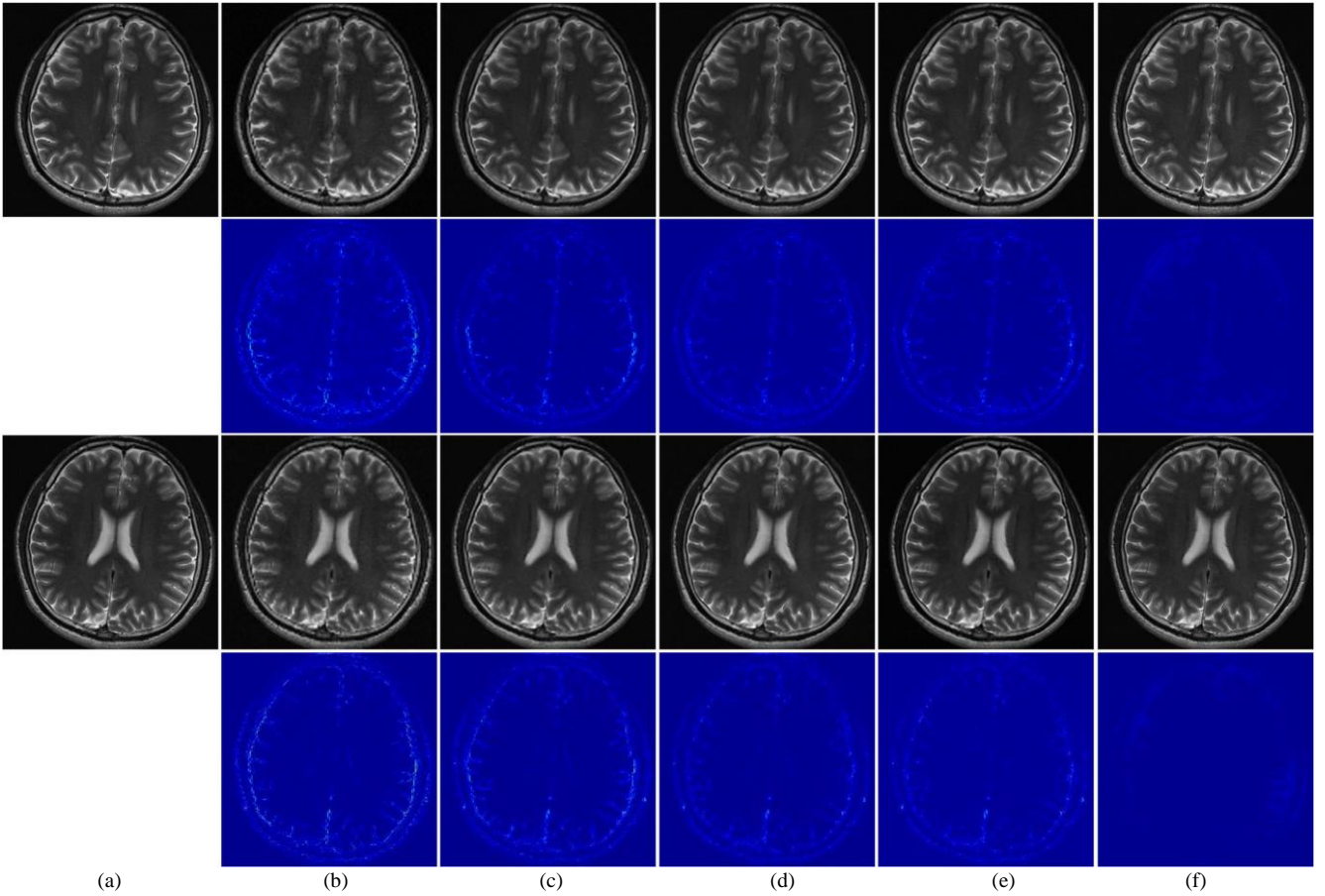

(a) (b) (c) (d) (e) (f)
**Fig. 7.** Comparison results of different algorithms on Brain dataset. (a) Reference image of Brain dataset; (b) BICUBIC; (c) SRCNN; (d) FAWDN; (e) SRGAN; (f) VCA-I; Bottom: The residuals between the reference images and reconstruction images.

### D. Ablation Study in K-space Domain

To evaluate the effectiveness of virtual coils technology, we carry out ablation studies. We use SCC [58] to generate 2-channel images from the original images. The dataset in k-space domain is obtained through the Fourier transform of image domain dataset. Taking Brain dataset as an example, 2-channel complex-valued images are used as the input of the network. At the same time, the data of 4-channel, 6-channel and 12-channel are used as labels to verify the effect of multi-channel property. It is noteworthy that the loss function of VCA-K is performed in image domain.

Table III tabulates the reconstructed PSNR and SSIM values of VCA-K can reach 42.47 dB and 0.9874 on Brain dataset. The 12-channel Brain dataset is 2.5 dB and 4.5 dB higher than the 6-channel Brain dataset and the 4-channel Brain dataset, separately. In Cardiac dataset, the PSNR of the reconstruction results increases after channel augmentation compared to the 2-channel dataset. Qualitatively, the same conclusion can be obtained from Fig. 8 and Fig. 9. It is consistent with the results obtained in image domain. Thus, it indicates that VCA-K not only has a better reconstruction effect in image domain as the number of channels increases, but also in k-space domain.

TABLE III
QUANTITATIVE RECONSTRUCTION (PSNR/SSIM) RESULTS FOR VCA-K ON BRAIN AND CARDIAC DATASETS.

| Brain | 2ch → 4ch | 2ch → 6ch | 2ch → 12ch |
|---|---|---|---|
| | 37.97/0.9761 | 39.93/0.9852 | 42.47/0.9874 |
| Cardiac | 2ch → 6ch | 2ch → 10ch | 2ch → 20ch |
| | 23.16/0.8259 | 23.15/0.8119 | 23.07/0.8157 |

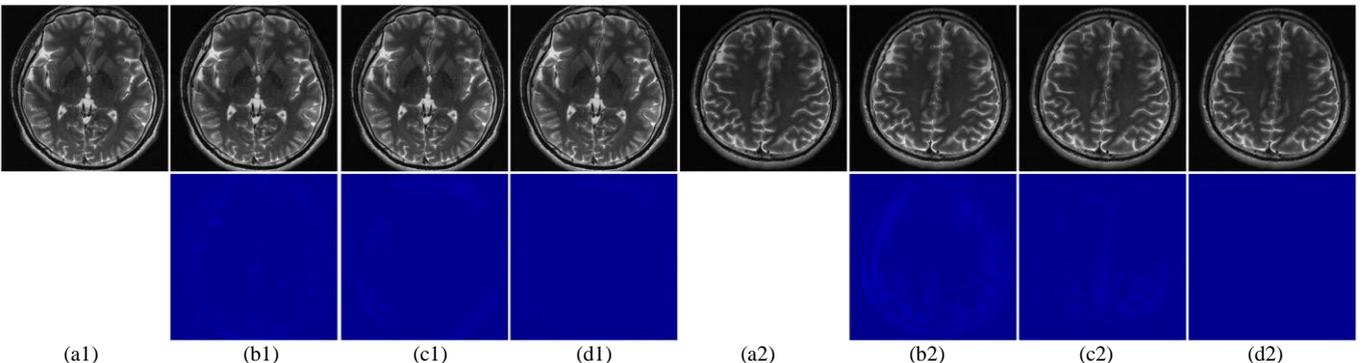

(a1) (b1) (c1) (d1) (a2) (b2) (c2) (d2)
**Fig. 8.** Reconstruction results of Brain dataset for VCA-K. Top: (a) Reference image of Brain dataset; Reconstruction results of (b) 2ch → 4ch; (c) 2ch → 6ch; (d) 2ch → 12ch; Bottom: The residuals between the reference images and reconstruction images.

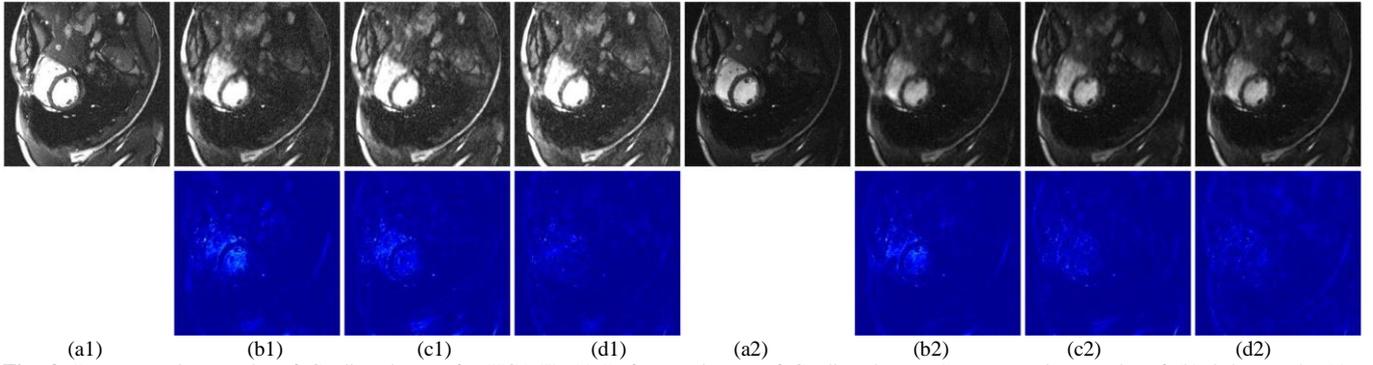
**Fig. 9.** Reconstruction results of Cardiac dataset for VCA-K. (a) Reference image of Cardiac dataset; Reconstruction results of (b) 2ch → 6ch; (c) 2ch → 10ch; (d) 2ch → 20ch; Bottom: The residuals between the reference images and reconstruction images.

*E. PI Comparison in K-space Domain*

To further evaluate the proposed VCA-K, we compare the reconstruction effects of PI algorithm before and after the channel expansion under different acceleration factors. We use three representative PI methods, which span various reconstruction classification models, including ESPIRiT [60], $L_1$ ESPIRIT [61] and AC_LORAKS [62]. The experiment is conducted under-sampling at equal intervals with acceleration factor of $R$ =2, 3, 4, and 5. The number of reference lines for auto-calibration and the calculation of the coil sensitivity profiles is 30. The data size employed to construct calibration matrix is $256 \times 30 \times 12$, and the kernel size (window size) is $6 \times 6$. Here, a lower threshold of 0.02 is used for calculation of the sensitivity maps to avoid truncation artifacts. We take ESPIRiT as an example to describe the reconstruction steps using PI algorithms after channel expansion. At the first step, virtual coils are formed for both ACS and under-sampled data according to variable augmentation technology. In the second step, a standard ESPIRiT reconstruction is performed and the resulting images are combined using the SOS operation.

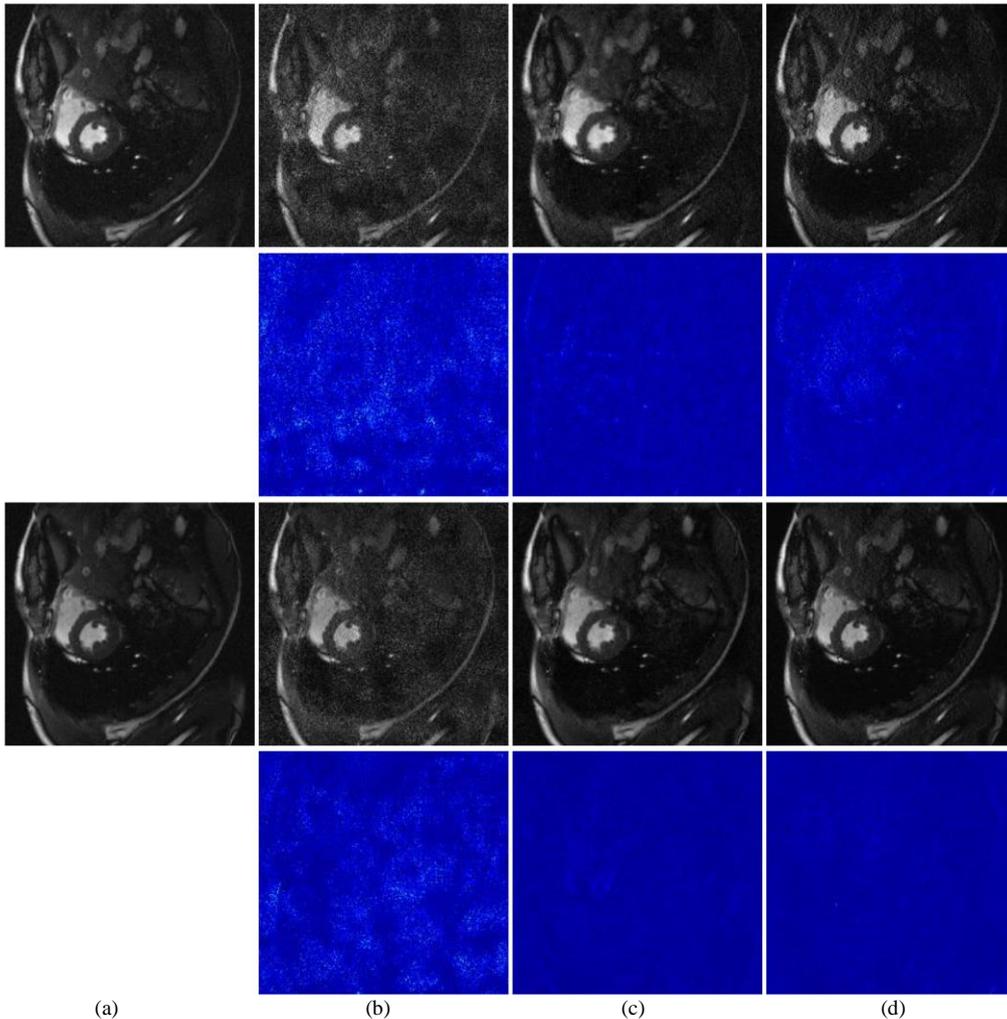
**Fig. 10.** Comparison results of the 20-channel object after channel expansion/extrapolation and the original 6-channel version at an acceleration factor of 4. (a) Reference image of Cardiac dataset; (b) ESPIRIT; (c) $L_1$ ESPIRIT; (d) AC_LORAKS; The first row is the original 6-channel data, and the third row is the reconstruction effect of the expanded 20-channel data. The second and fourth rows are residuals between the reference images and reconstruction images.

We use the high-dimensional information formed by channel expansion as the prior information of PI to improve the reconstruction effect. The result before channel expansion is calculated using 6-channel of under-sampled data as input and 6-channel of fully sampled data as label. The result after channel expansion is calculated by using the expanded 20-channel under-sampled data as the input and using the 20-channel full-sampled data as the label. Comparisons of image reconstruction performance under different sampling rates for Cardiac dataset are summarized in Table IV. Specifically, the PSNR and SSIM values of the three algorithms are improved at different sampling rates. Since the k-space is variable density randomly under-sampled, it tends to produce structure-like artifacts. Visually, the performances of reconstruction for different methods are various. It is worth noting that the three reconstruction methods are significantly improved after channel expansion. It can also be seen from the residual that the reconstruction effect after channel replication is better than the previous reconstruction effect. Extra information provided by the virtual coils is available for synthesizing the target signal, thereby improving the reconstruction conditions and image quality. It further indicates that VCA-K can not only improve the image reconstruction quality in the case of full sampling, but also contribute to the further reconstruction of the PI algorithm in the case of under-sampling.

TABLE IV
COMPARISON RESULTS OF 20-CHANNEL AFTER CHANNEL EXPANSION AND THE ORIGINAL 6-CHANNEL WITH THREE STATE-OF-THE-ART METHODS IN TERMS OF PSNR AND SSIM.

| Dataset | | ESPIRIT | $L_1$ ESPIRIT | AC_LORAKS |
|---|---|---|---|---|
| $R=2$ | 6 coils | 21.92/0.4320 | 30.68/0.7706 | 32.15/0.8396 |
| | 6 coils→20 coils | 25.60/0.5707 | 33.44/0.8515 | 35.67/0.9148 |
| $R=3$ | 6 coils | 21.22/0.3845 | 29.05/0.7008 | 28.68/0.7246 |
| | 6 coils→20 coils | 23.69/0.4895 | 32.66/0.8306 | 32.95/0.8605 |
| $R=4$ | 6 coils | 21.65/0.3732 | 28.16/0.6790 | 27.01/0.6483 |
| | 6coils→20 coils | 24.30/0.5066 | 31.74/0.8043 | 31.16/0.8080 |
| $R=5$ | 6 coils | 22.23/0.3877 | 27.70/0.6648 | 24.90/0.5742 |
| | 6 coils→20 coils | 24.18/0.4886 | 31.06/0.7900 | 30.02/0.7686 |

## V. CONCLUSIONS

In this paper, a method of using artificial intelligence to expand the coil arrays in MRI was proposed. A reversible network was used as the central part of the training network, and auxiliary variable technology was used to make the number of channels output by the network equal to the input of the network. Dummy variable technology was utilized to expand the channel/coil to achieve the purpose of increasing the virtual coil arrays. The present VCA principle can be conducted in both image domain and k-space domain. i.e., VCA-I expanded the channel in image domain, and VCA-K expanded the channel in k-space domain. The main advantage of VCA-K was the easy implementation for improved PI performance without requiring modifications of the existing algorithms, thus paved a new way to leverage the PI. Comparative experiments demonstrated that VCA could obtain more clear and trusted medical images than state-of-the-art methods.